\documentclass[sigconf,screen]{acmart}
\usepackage{multicol}
\usepackage{multirow}
\usepackage{colortbl}
\usepackage{subcaption}
\usepackage{hyperref}

\AtBeginDocument{%
  \providecommand\BibTeX{{%
    \normalfont B\kern-0.5em{\scshape i\kern-0.25em b}\kern-0.8em\TeX}}}

\copyrightyear{2021}
\acmYear{2021}
\setcopyright{none}

\acmBooktitle{4th ACM SIGSPATIAL International Workshop on AI for Geographic
Knowledge Discovery (GEOAI '21), November 2--5, 2021, Beijing, China}
\acmPrice{15.00}
\acmDOI{10.1145/3486635.3491070}
\acmISBN{978-1-4503-9120-7/21/11}

\begin{document}

\title{Synthetic Map Generation to Provide Unlimited Training Data for Historical Map Text Detection}

\author{Zekun Li}
\email{li002666@umn.edu}
\affiliation{%
  \institution{University of Minnesota}
  \city{Minneapolis}
  \country{USA}
}

\author{Runyu Guan}
\email{guanyu@usc.edu }
\affiliation{%
  \institution{University of Southern California}
  \city{Los Angeles}
  \country{USA}
}
\author{Qianmu Yu}
\email{qianmuyu@usc.edu }
\affiliation{%
  \institution{University of Southern California}
  \city{Los Angeles,}
  \country{USA}
}
\author{Yao-Yi Chiang}
\email{yaoyi@umn.edu}
\affiliation{%
  \institution{University of Minnesota}
  \city{Minneapolis}
  \country{USA}
}
\author{Craig A. Knoblock}
\email{knoblock@isi.edu}
\affiliation{%
  \institution{University of Southern California}
  \city{Los Angeles}
  \country{USA}
}

\renewcommand{\shortauthors}{Zekun Li et al.}

\begin{abstract}
 Many historical map sheets are publicly available for studies that require long-term historical geographic data. The cartographic design of these maps includes a combination of map symbols and text labels. Automatically reading text labels from map images could greatly speed up the map interpretation and helps generate rich metadata describing the map content. Many text detection algorithms have been proposed to locate text regions in map images automatically, but most of the algorithms are trained on \textit{out-of-domain} datasets (e.g., scenic images). Training data determines the quality of machine learning models, and manually annotating text regions in map images is labor-extensive and time-consuming. On the other hand, existing geographic data sources, such as OpenStreetMap (OSM), contain machine-readable map layers, which allow us to separate out the text layer and obtain text label annotations easily. However, the cartographic styles between OSM map tiles and historical maps are significantly different.  This paper proposes a method to automatically generate an \textit{unlimited} amount of annotated historical map images for training text detection models. We use a style transfer model to convert contemporary map images into historical style and place text labels upon them. We show that the state-of-the-art text detection models (e.g., PSENet) can benefit from the synthetic historical maps and achieve significant improvement for historical map text detection.

\end{abstract}


\maketitle

\section{Introduction}

Historical maps are excellent sources for understanding human activities and city development~\cite{Chiang2020-vm}. Many organizations such as the United States Geological Survey (USGS)\cite{usgs}, Esri\cite{esri}, and National Library of Scotland (NLS) \cite{odmaps} have made a great effort in scanning historical maps and releasing them for public use. US Fire Insurance Atlase \cite{sanborn} digitized by the New York Public Library (NYPL) \cite{nypl} documents the change of environment and geography of the New York City during the 19-th and early 20-th centuries. The USGS topographic maps \cite{usgs} preserve the past landscape of the entire country during the 19-th century and provides invaluable support for physical and cultural studies. The Ordnance Survey \cite{odmaps} publishes a variety of large-scale maps that cover the Great Britain area in the 19-th century. These existing map series were initially created for different reasons, such as taxation, and have greatly served the purpose in that era. Nowadays, they continue to offer a channel for us to look back in time. 


\begin{figure}
\includegraphics[width=\linewidth]{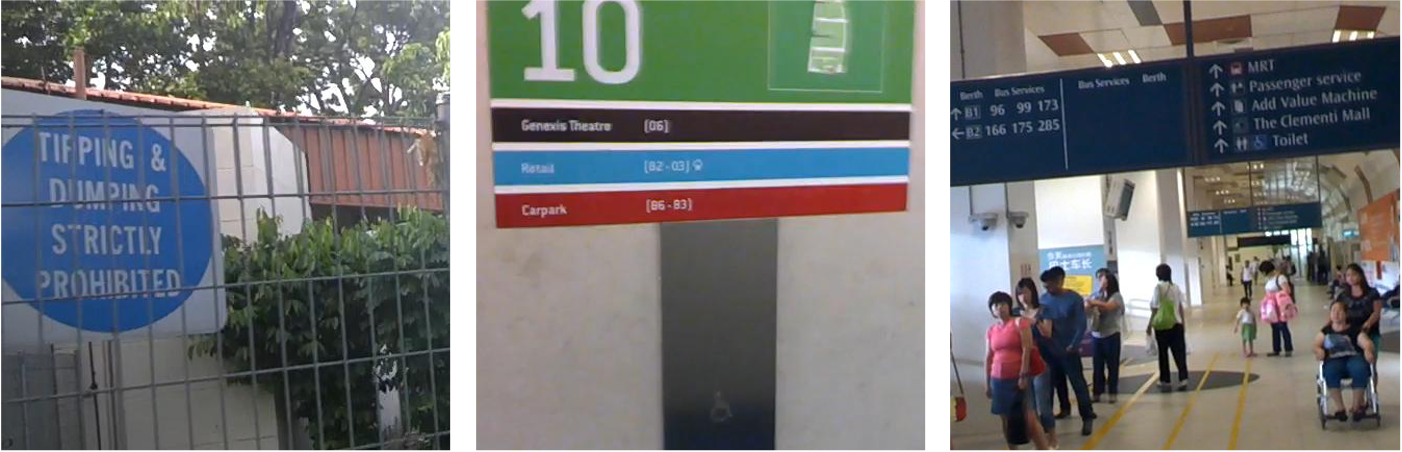}
\includegraphics[width=\linewidth]{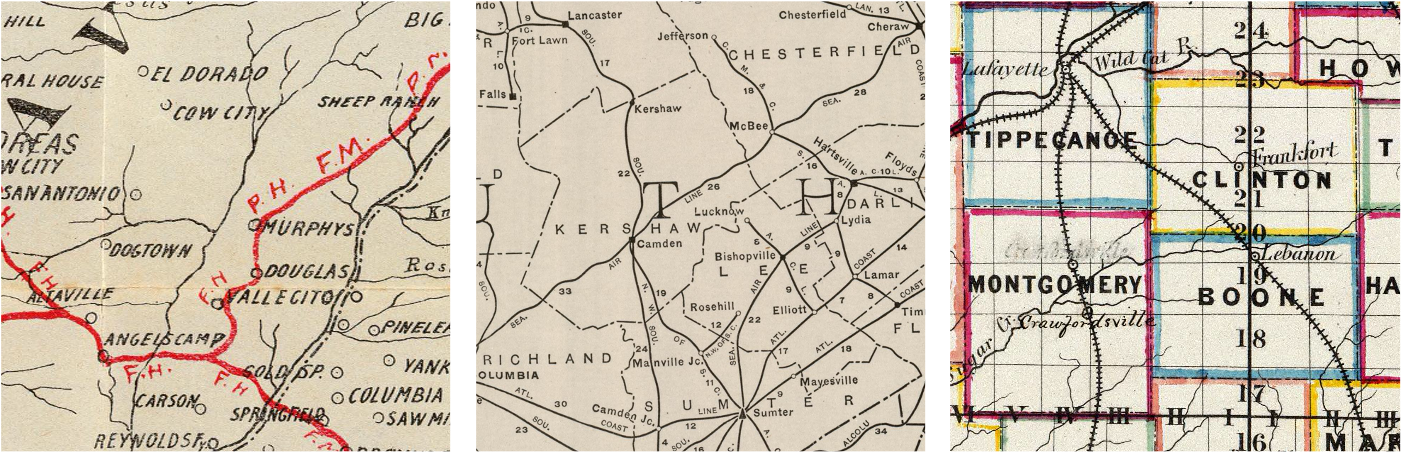}
\caption{\textit{First Row}: Sample images from the ICDAR2015 Incidental Scene Text dataset. \textit{Second Row}: Sample image patches from the David Rumsey Maps dataset. Text labels in these two datasets differ significantly in text arrangement and style.  }
\label{fig:sample_images}
\end{figure}


In the recent years, researchers have attempted to automatically extract text labels and produce metadata from historical maps~\cite{li2020automatic,pezeshk2011automatic}. In the mean time, many deep learning approaches have been developed for detecting text in electronic documents or scene images~\cite{wu2017self,jiang2017r2cnn,zhou2017east}. All these methods need to be trained with a lot of training data to obtain the best performance. Fortunately, the International Conference on Document Analysis and Recognition (ICDAR) released several datasets to address text detection problems for scene image detection and scanned documents. For both historical map text detection and scene text detection, there are some challenges that are common in both domains. For example, the variance of the font size can be large in both map and scene images. Also, images in both domains use many different font styles in the documents. However, some characteristics are \textit{unique} to historical maps. Historical maps often have a noisy (high edge intensity) background, such as complex road networks or contour lines in mountainous areas, while other electronic documents and scene images usually have a simpler, homogeneous background within the text region. Some map text labels (\textit{e.g.,} street names) can have large spacing between characters. Moreover, map text can be oriented and curved to follow the given underlying geographic features, such as railroads, boundary lines, and rivers. In contrast, documents and scene images usually have straight text in the horizontal or vertical directions. These differences pose new challenges to the text detection algorithms for handling historical maps. Due to these differences, text detection models trained on scene images may not perform well on historical map images.

To adapt existing text detection models to the historical map domain, we need to feed the model some training data in the historical map domain. However, the labeled training data does not come for free. Work is required to draw the bounding boxes/polygons around the text regions, which can be quite time-consuming. This paper proposes a method to automatically generate a large amount of training data for the historical map text detection with minimal manual work. 

The general idea is that, we first produce a synthetic historical map background layer without any text labels and then automatically place text labels upon the layer. Since we have full control over the text layer, ground-truth (annotation) information (i.e., text bounding polygon) can be recorded automatically. Specifically, we use a \textit{style transfer} model CycleGAN to convert OpenStreetMap raster images to the historical style and then use the QGIS PAL API \cite{pal} to place the text labels on the historical map background. The QGIS PAL API is able to place text labels according to the position and geometry of the underlying geographical feature. For Point features, the API places the label around the point. For Line features, the API places the label along the line. After generating the synthetic historical map background and placing the text labels, we also have a method to compute the ground-truth bounding polygons from the synthetic map image. We use the bounding polygon instead of bounding rectangle representation, because the text labels can be curved and sometimes arbitrarily shaped. The rotated bounding rectangles are not tight enough to enclose the text region accurately. 

The main contribution of the proposed approach is an end-to-end pipeline to generate a large amount of annotated training data, enabling the use of deep learning models for unlocking useful textual information from historical map images. There are three major advantages of the proposed approach: (1) Once the style transfer model is trained on one map style, it can then generate an unlimited number of images in this style. The dataset size is guaranteed to be sufficient to train the deep-learning text detection models. (2) The CycleGAN style transfer model does not need paired data for training. Hence, the historical map images do not need to cover the same region as the OSM data. (3) The style transfer model can produce synthetic historical map images with any style, as long as a small amount of training data is provided to initialize the style transfer. No labeling information is required in the end-to-end process. In the experiment section, We show that the PSE-Net, a deep-learning based text detection model, can achieve improved performance after fine-tuning it on the proposed synthetic map dataset.

\begin{figure*}
\includegraphics[width=\textwidth]{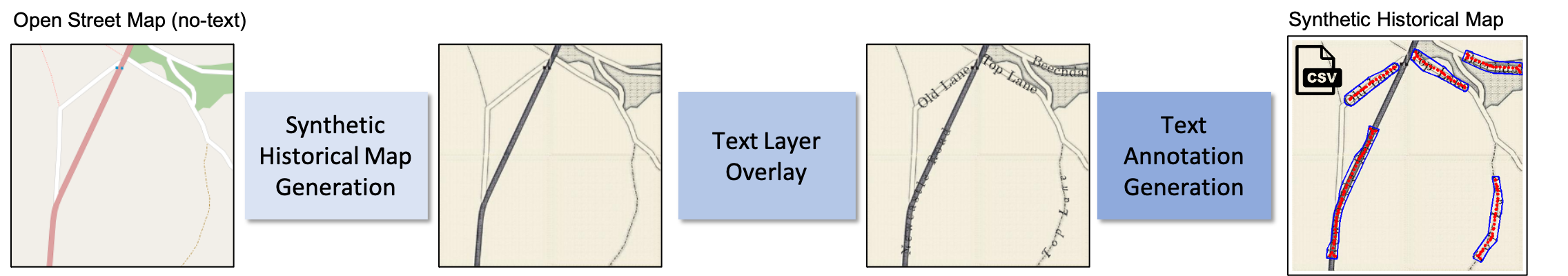}
\caption{Pipeline to generate a large amount of training data for text detection on historical maps. We first use a style transfer network CycleGAN to convert an OSM image to the historical style, then associate the font, style and placement strategy according to the underlying geographical feature. We use QGIS PAL API to place the text labels on the synthetic map background, and design an approach to automatically generate the polygon, centerline and local height annotation for the text labels. }
\label{fig:outline}
\end{figure*}

\section{Approach}
In this section, we first describe the two datasets that we use to provide the \textit{source} and \textit{target} style, then explain the synthetic historical map generation process in detail. There are three main steps involved: synthetic historical map generation, text layer overlay, and text annotation generation. The source code and the dataset that we use to train the model is available at \hypertarget{https:\/\/github.com\/zekun-li\/generate_synthetic_historical_maps}{https://github.com/zekun-li/generate\_synthetic\_historical\_maps}, and a live demo to show the style-transferred synthetic historical map is available at \hypertarget{https:\/\/zekun-li.github.io\/side-by-side\/}{https://zekun-li.github.io/side-by-side/}.

\subsection{Data Sources for Style Transfer}
We employ two data sources for the synthetic historical map generation. (1) \textit{Open Street Maps} (OSM) data, which provides the source image for style transfer. (2) \textit{Ordnance Survey} 6-inch maps in the years of 1888-1913 (also referred to as the GB1900 6-inch layer on the National Library of Scotland website\footnote{https://geo.nls.uk/maps/gb1900/}). The OSM data provides the content of the synthetic map, and the Ordnance Survey data provides the historical style of the synthetic map. We choose OSM as the source image dataset because it is an open-source dataset with data coverage over the full globe. It is easy to obtain both the vector data and rasterized image tiles from the OSM. We use the Ordnance Survey map sheets for target historical style since the 6-inch map covers the whole Britain area, and all the map sheets have been georeferenced by the National Library of Scotland. 

\begin{figure}[t]
  \centering
  \includegraphics[width=0.8\linewidth]{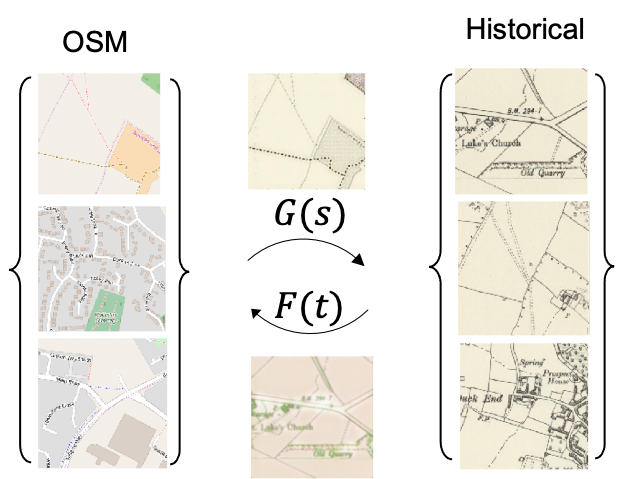}
  \caption{Illustration of CycleGAN. Generator $G$ learns to convert OSM images to historical style and another generator $F$ learns to convert the historical map images to the OSM style. During training, both generators are trained together with domain-specific discriminators. During synthetic data generation, only $G$ will be used to synthesize historical map images from OSM. }
  \label{fig:cyclegan}
\end{figure}

\subsubsection{Open Street Map (OSM)}
There are two groups of data involved from OSM. One is the data used to train the CycleGAN style transfer model, and the other is the data used to generate a large amount of historical map images for the downstream tasks. There is no limitation of whether these two groups of data should be in the same region or not. In our experiments, we used different regions. Group 1 data is randomly downloaded from the Great Britain region, and the second group is around the Birmingham region. We downloaded the data at zoom level 16, and this yields 27,707 tiles with size 256x256 in group 1 and 54,865 tiles in group 2. The raster data for group one and two is downloaded from the WMFLabs tile server\footnote{https://tiles.wmflabs.org/osm-no-labels/\$\{z\}/\$\{x\}/\$\{y\}.png} which does not contain text labels. The vector data for group two is downloaded from the Geofabrik website.\footnote{https://download.geofabrik.de/europe/great-britain/england.html} There is no need for vector data for training the CycleGAN model. 

\subsubsection{Ordnance Survey Historical Maps}
While adding the text layers, we prefer to convert OSM raster images to the historical map style and use the synthetic map as the map ground,  instead of using the real historical map directly as the background, because there is no easy way to accurately remove the text labels from the existing historical map images. Since removing the text labels requires the knowledge of text location in advance (a.k.a text detection) and this leads to the chicken and egg problem. 

We only need the Ordnance Survey historical map data for training the CycleGAN model, and do not need real historical map anymore when the synthetic map images have been produced. In terms of the study area, we used the same region as OSM group one although there is no requirement of these two data sources need to cover the same area. The raster tiles are also retrieved at zoom level 16 with size 256x256.

\begin{figure*}[t]
  \centering
  \includegraphics[width=\linewidth]{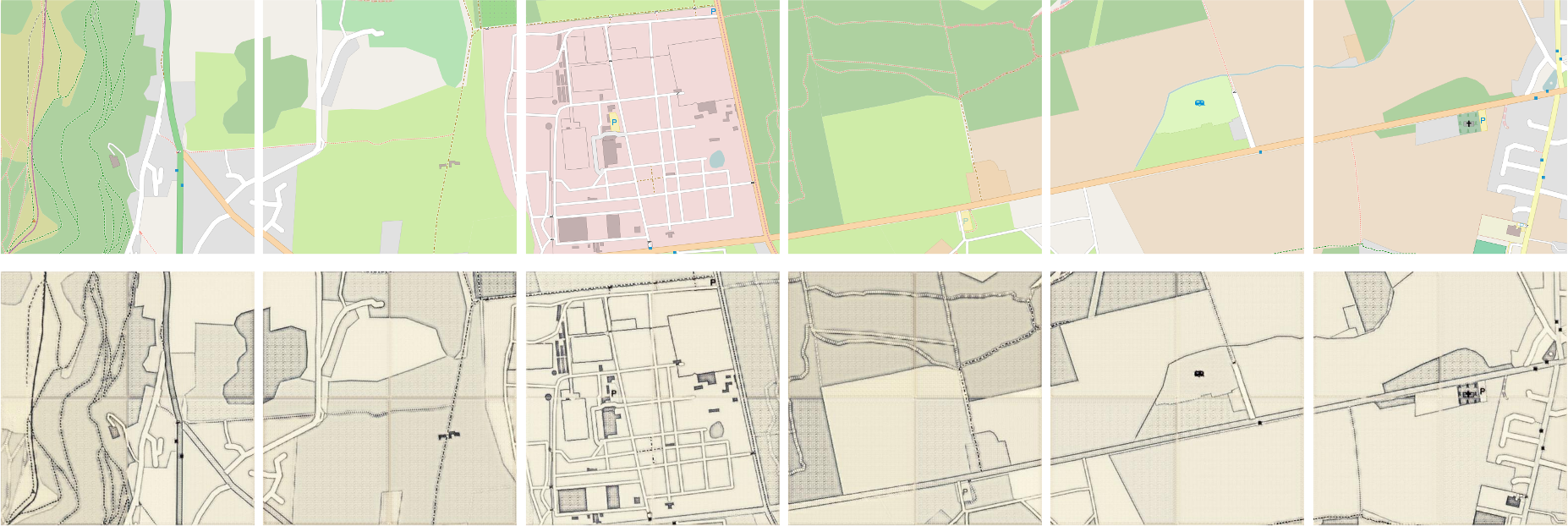}
  \caption{Visualization of the Open Street Map (OSM) tiles and the output synthetic historical map tiles.}
  \label{fig:osm_syn}
\end{figure*}

\subsection{Synthetic Historical Map Generation}

The idea of style transfer is built upon the Generative Adversarial Network (GAN). GAN models have a discriminator and a generator. The generator is responsible for generating fake images, while the discriminator tries to distinguish fake images from real images. The two modules keep combating each other, and the discriminator improves its ability to tell real from fake, and the generator keeps generating images with better and better quality. 

The difference between CycleGAN and other GAN models is that the cycleGAN has two generators and discriminators. The two generators are used to generate images with the two given styles, and the discriminators are used to distinguish the images for two styles, respectively. Hence, the network can convert the images with style $S$ to style $T$ and then convert them back to $S$.

Formally, we can define the process as following. Let $S=\{s_i\}_{i=1}^M$ be the set of $M$ Open Street Map images which do not contain any text labels, and $T=\{t_i\}_{i=1}^N$ be the set of $N$ historical map images. We define a Generator $G:S\rightarrow T$ that learns to translate $s_i$ to $t_i$. Also, we define another generator $F: T\rightarrow S$ that translate $t_i$ back to $s_i$. The \textbf{\textit{Cycle Consistency Loss}} defined in Eq. \ref{eq:loss-cycle} encourages $G(F(T)) \approx T$ and $F(G(S)) \approx S$. Meaningly, if an image is fed through both generators sequentially, the output image should look very similar to the original image itself with $x_i \approx G(F(x_i))$ and $x_i \approx F(G(x_i))$. 

\begin{equation}
\label{eq:loss-cycle}
\begin{split}
\mathcal{L}_{cycle}(G,F) =& \mathbf{E}_{s\sim p(s)}[||F(G(s))-s||_1] \\
    +& \mathbf{E}_{t\sim q(t)}[||G(F(t))-t||_1] 
\end{split}
\end{equation}

To ensure the high-quality of generated images, two discriminators designed for each style are employed to distinguish the real images from generated ones using \textbf{\textit{Adversarial Loss}} in Eq. \ref{eq:loss-adv}. Specifically, we have $D_S$ and $D_T$, where $D_S$ tries to discriminate between original images in $S$ and the generated images in $F(T)$, and $D_T$ tries to discriminate between original images in $T$ and the generated images in $G(S)$. 

\begin{equation}
\label{eq:loss-adv}
\begin{split}
\mathcal{L}_{adv}(G,F,D_S,D_T) =& \mathbf{E}_{t\sim q(t)}[log D_T(t)] \\
    +& \mathbf{E}_{s\sim p(s)}[log (1-D_T(G(s))] \\
    +& \mathbf{E}_{s\sim p(s)}[log D_S(s)] \\
    +& \mathbf{E}_{t\sim q(t)}[log (1-D_S(F(t))] \\
\end{split}
\end{equation}

In summary, the total loss $\mathcal{L}(G,F,D_S,D_T)$ is composed of two parts: the cycle-consistency loss $\mathcal{L}_{cycle}$ and the adversarial loss $\mathcal{L}_{adv}$, and it can be written as $\mathcal{L}(G,F,D_S,D_T) = \mathcal{L}_{cycle}(G,H) + \lambda \mathcal{L}_{adv}(G,F,D_S,D_T)$.

To generate the synthetic historical map background images, we take the trained model $G$ and feed $S$ as input. The output is a set of images $T' = G(S) = \{t'_i\}_{i=1}^M$ from OSM dataset whose style has been translated into historical style. Figure \ref{fig:osm_syn} shows some sample images from OSM and the output synthesized map tiles. 

\subsection{Text Layer Overlay}
\subsubsection{Font Size and Style} According to the underlying geographical feature type, we roughly divide the font size into three levels: Large, Medium, and Small. The large labels correspond to the geographical features covering very large regions, and small ones correspond to smaller regions. For the font style, we use several fonts downloaded from FontSpace\footnote{https://www.fontspace.com/category/antique} and the Cheysson font from the ArcGIS website\footnote{https://www.arcgis.com/home/item.html?id=6b12e5149fd549f4829725ea46affb55}. We also include several MacOS system fonts in the font family, which makes a total of 16 fonts. Each geographical feature type has an associated font style and size, and the text labels with the same underlying geo-feature have the same font style and size. Table \ref{tab:font_size} shows the statistics of the font size information.

\begin{table}
  \caption{Font Size Statistics}
  \label{tab:font_size}
  \begin{tabular}{ccl}
    \toprule
    Groups & Size (pt) & Geo Features  \\
    \midrule
    Large &    [60,80] & canal, city, county, town, village \\
    & &  waterfall, wetland, island \\
    Med. &  [35,45] & airfield, airport, allotment, archaeological\\
    & & battlefield, camp site, cliff, dock, farmland\\
    & & farm, forest, fort, hamlet, nature reserve \\
    & & reservoir, ruins, vineyard, rail river, stream \\
    Small  & [20,30] & others (e.g. streets) \\
  \bottomrule
\end{tabular}
\end{table}
\subsubsection{Text Label Placement}
\begin{figure}[t]
  \centering
  \includegraphics[width=\linewidth]{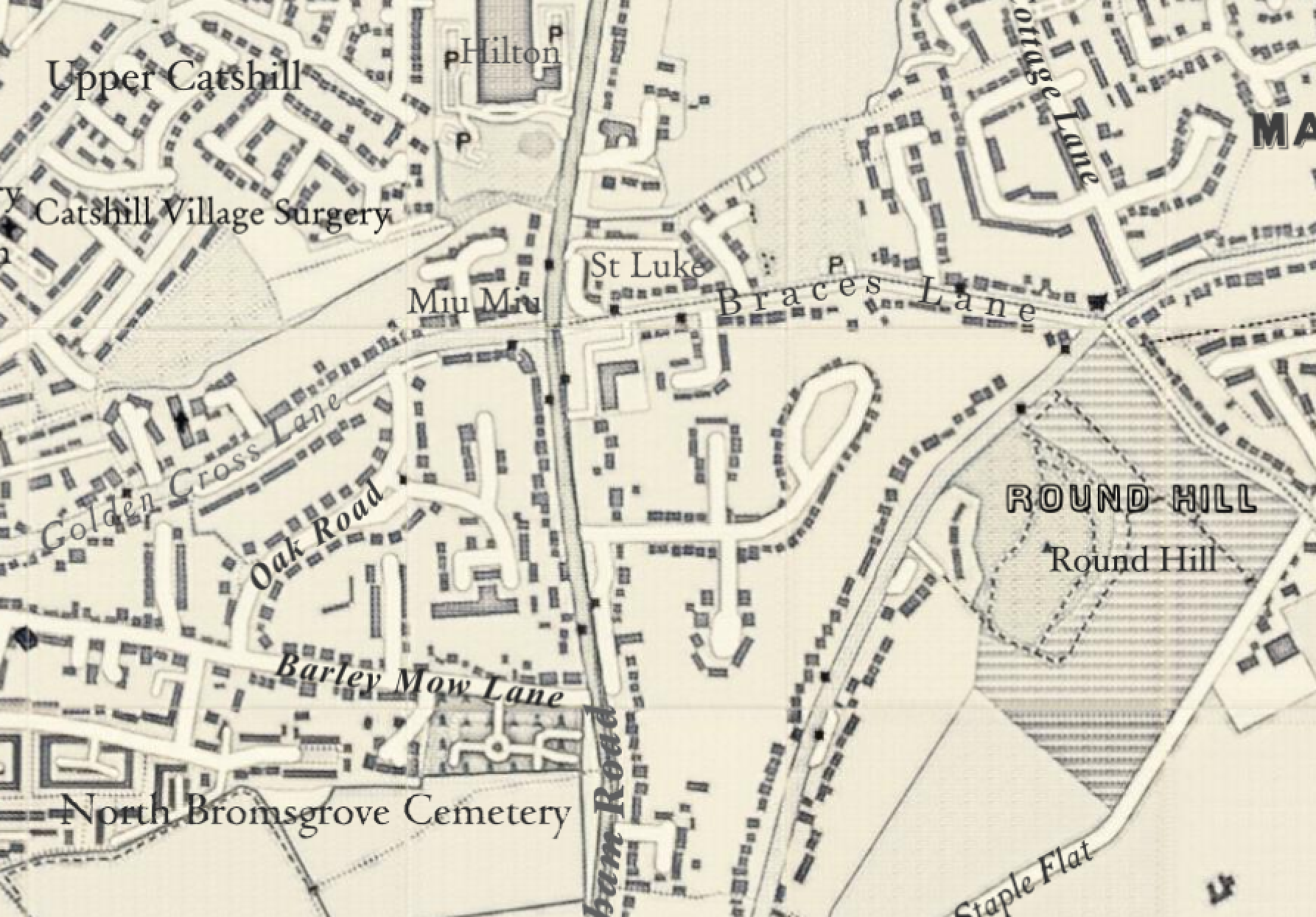}
  \caption{Sample synthetic map region generated by our model. Source map is Open Street Map (OSM) and the target style is the Ordnance Survey 6-inch historical map. Text labels come from the vector data of OSM.}
  \label{fig:synthmap}
\end{figure}

We utilize the QGIS PAL API for text label placement. For Point features, the text labels are placed around the point. For the MultiLine geo features, the text labels are placed on the center of the line. For MultiPolygon geofeatures, labels are placed around the center of the area. There are no overlapping or intersecting text labels for any of the geo features. Specifically, the underlying geo features might overlap with other features or text labels, but the text labels should not overlap with each other. Figure~\ref{fig:synthmap} shows a sample map region after the text labels are placed on the synthetic map.

\subsection{Text Annotation Generation}
We provide two representations of the text annotations: (1) Bounding polygons - a tight concave polygon for each text label (2) Centerlines and local heights - the centerline is provided in the form of a sequence of points, and local height can be thought of as the height of the bounding polygon. A bounding polygon can be constructed when centerline and local height are both known. The reason we provide two types of annotation is that some text detection algorithms (e.g., TextSnake\cite{long2018textsnake}) are centerline and local height-based, while some others are polygon-based (e.g., PSENet \cite{wang2019shape}).

\subsubsection{Bounding Polygon}\label{sec:bounding_polygon}
When rendering the text layer with QGIS, we produce two versions of the raster image: the colored version and the gray-scale version. We set the non-text region to be transparent for both versions and keep the text labels at exactly the same position. In the colored version, each location name label is painted with a different color. Thus it is easy to 1) separate all the text labels from the transparent non-text region and 2) separate one particular text label from all other labels. We convert the colored version from RGBA space to the Black/White (BW) space to produce the gray-scale version. The gray-scale version is then added to the synthetic historical map background to render the complete map. 

By differentiating the color of the pixels from the colored version, we can obtain all the pixels belonging to the same text label. We call the text region pixels as the foreground and other unrelated pixels as the background. We then filter out the background color to obtain the positions of the foreground pixels. Finally, we compute the concave hull of the foreground pixels to generate the final bounding polygon. We adopt the \textit{alphashape} algorithm for the concave hull computation and set the $\alpha$ parameter to be 0.02 empirically for all the text labels. Following the ICDAR datasets convention, we store the polygon points in clockwise order.

\subsubsection{Centerline and Local Height}
The centerline and local height representation offer another way to describe the ground truth. The centerline is a multi-segment line across the centerline pixels of the text region. The local height denotes the height $h$ (or diameter) of the text region.

\begin{figure}[h]
  \centering
  \includegraphics[width=\linewidth]{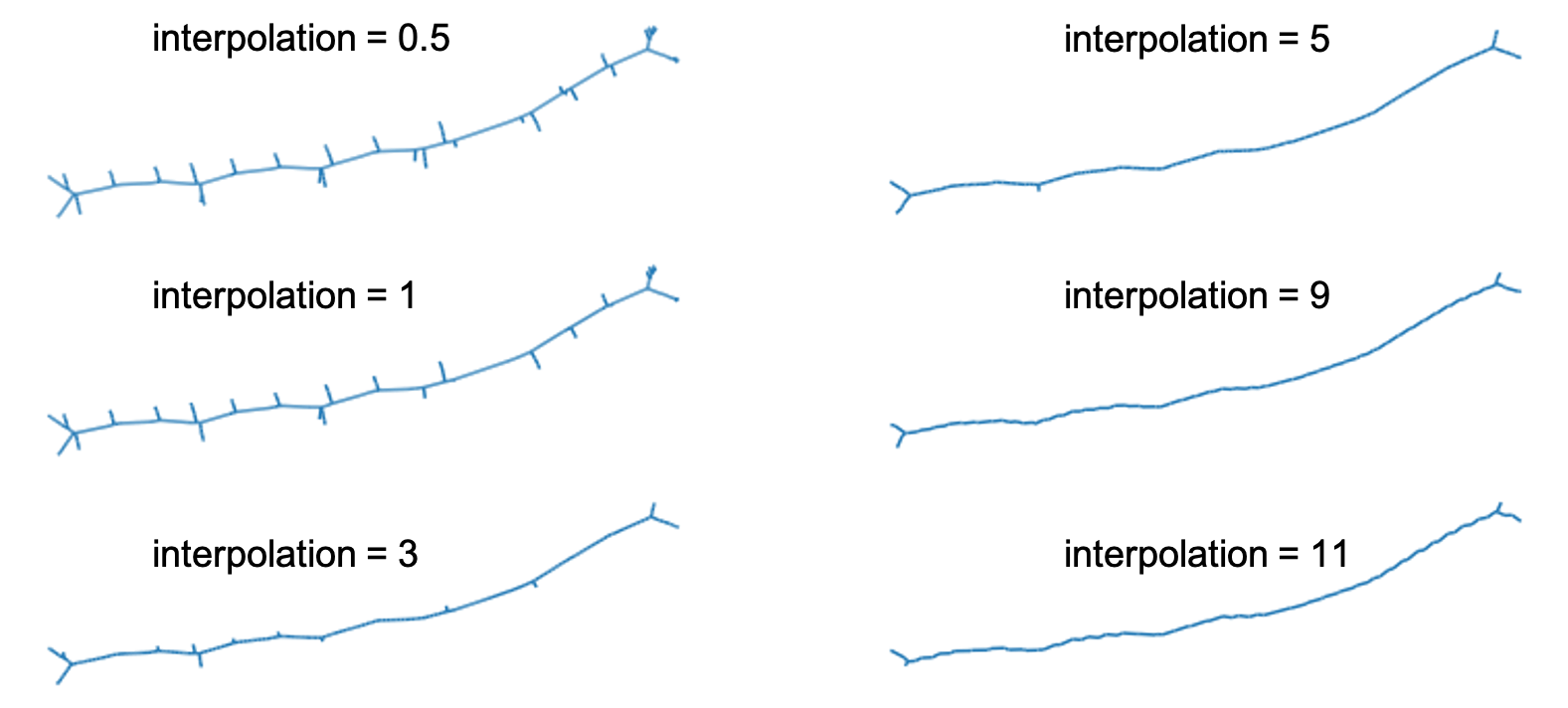}
  \caption{Centerline computed with different interpolation distances. Larger interpolation values lead to smoother centerline.}
  \label{fig:centerline}
\end{figure}

For the centerline computation, we use an existing Python package called \verb|centerline,|\footnote{https://pypi.org/project/centerline/} which utilizes the Voronoi diagram to compute the centerline for the polygon. The border density parameter controls how many points to sample inside the polygon. With small border density values, the resulting centerline will contain a lot of details and is likely to form a tree structure as shown in Figure~\ref{fig:centerline}. Larger border density values lead to a smoother centerline. In our experiments, we empirically set the border density parameter (interpolation distance) to be 9. But even with a large border density, there are still some branching lines at the two ends of the centerline, as shown in Figure \ref{fig:centerline}.

\begin{figure}[t]
  \centering
  \includegraphics[width=0.7\linewidth]{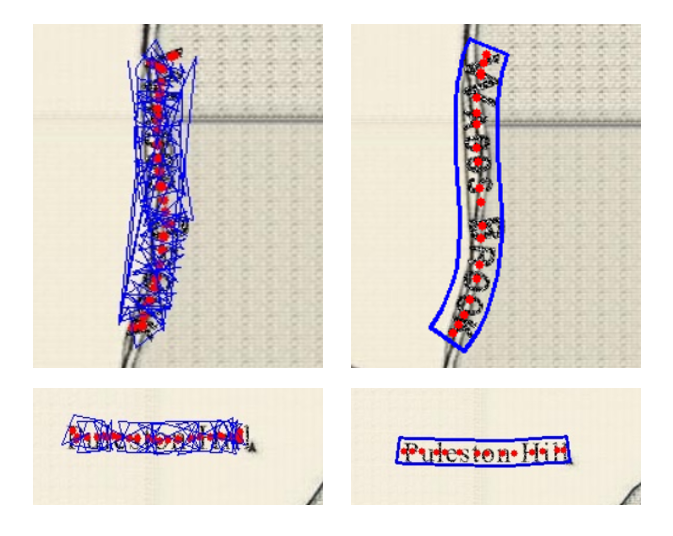}
  \caption{Centerline-based bounding polygon produced with original centerline with branches (left) and neat centerline (right). The red points are the centerline locations and blue line segments are the edges in the bounding polygon. The left ones have messy polygons because the centerline points are not sorted sequentially. }
  \label{fig:abp_cent}
\end{figure}

To further avoid tree branches and make the centerline generation robust to different interpolation values, we use the cubic curve fitting function to generate neat centerlines that do not branch at any point. For curve fitting, it is common to have the $x$-axis (horizontal) values as the independent variables and fit a curve $y=f(x)$. However, if the centerline is almost vertical, the result of using $x$ values as independent variables will be poor (see ``Wacos Brook'' in Figure \ref{fig:abp_cent}). Instead, for this case, we should fit along the $y$-axis by $x=f(y)$. To determine which axis values to use as the independent variable, we use a simple condition checking that computes the range of the $x$-axis values and $y$-axis values. Specifically, we first calculate the maximum and minimum of the $x$-axis values and $y$-axis values of the original centerline points: $X_{max}$, $Y_{max}$, $X_{min}$ and $Y_{min}$. Then we obtain the range of $x$-axis and $y$-axis with: $X_{var} = X_{max} - X_{min}, Y_{var} = Y_{max} - Y_{min}$. We choose the axis with a larger range as the independent variable. The second image in Figure \ref{fig:abp_cent} shows the final fitted centerline.

For the local height computation, we design a distance-transform based algorithm to determine the height of the text region. Similar to Section \ref{sec:bounding_polygon}, we first use pixel color information to get an image patch with only one text label. Given this color image patch $I$ and the polygon $P$ computed from \ref{sec:bounding_polygon}, we binarize $I$ to generate a masked version of the image $M$ where pixels inside the polygon are assigned to 1 and 0 otherwise. Let $F = \{M_{i,j} = 1\}$ be the set of foreground pixels and $B = \{M_{i,j} = 0\}$  be the set of background pixels. We then compute the Euclidean distance from each foreground pixel to the background pixel and let the maximum distance be the local height of the text region. 
\begin{align}
    h = max(||F_i, B_j||_2) && \forall i\in F, j \in B
\end{align}

\section{Experiments and Analysis}

\subsection{Datasets}

\subsubsection{ICDAR 2015 Incidental Scene Text}
International Conference on Document Analysis and Recognition (ICDAR) started releasing datasets for text detection and recognition in 2011. The training set in Incidental Scene Text contains 1,000 images with about 4,500 words. The text regions are annotated with tight quadrangles. This dataset has large variances in text font sizes, styles, and perspective angles, and mainly focuses on detecting text regions on scenic images. Some sample images from this dataset are shown in Figure \ref{fig:sample_images}. This dataset is used to train the PSENet in the first setting. 

\subsubsection{SynthMap}
This is the dataset that we introduce in this paper. It contains 13,892 synthetic map tiles style transferred from the OSM map images. Each map is of size 512x512 (concatenated from the 256x256 tiles), and the number of text regions on each map varies depending on the density of the underlying geographical feature. There are 45,375 text regions in total. The annotation information contains bounding polygon, centerline points, and the local height of the text region.

\subsubsection{David Rumsey Maps}
Weinman \cite{weinman2017geographic} collected 31 historical maps in the North America area from the David Rumsey map collection\footnote{http://davidrumsey.com} which span from 1866 to 1927. The map contains 12,578 words with 9,555 phrases. The map image has been manually annotated with quadrangles. The map images contain 9 series, and the number of maps from each series is listed in Table \ref{tab:weinman_dataset}. 

\begin{table}
  \caption{David Rumsey Maps Statistics (SID: Series ID, map images in the same series have similar style, \# Maps: number of map images in that map series. \# Text: average number of text regions in each map sheet)}
  \label{tab:weinman_dataset}
  \begin{tabular}{cccl}
    \toprule
     SID & \# Maps & \# Text & Study Areas\\
    \midrule
     D0006 & 1 & 553 & Tennessee\\
    D0017 & 1 & 653 & Stanislaus\\
    D0041 & 2 & 485 & Florida\\
    D0042 & 12 & 786 & Ohio, New Mexico, Indiana, Illinois\\ 
    &&& Michigan, Wisconsin, Minnesota, Iowa\\
    &&& Missouri, Kansas, Arkansas \\
    &&&  Mississippi, Alabama,  Nebraska\\
    D0079 & 1 & 354 & US \\
    D0089 & 1 & 671 & Northern Pacific\\
    D0090 & 1 & 607  & Missouri\\
    D0117 & 6 & 1902 & Indiana, Iowa, Nebraska, Colorado, \\
    &&& Wyoming, Montana\\
    D5005 & 6 & 1534& North Carolina, South Carolina\\
    &&& Minneapolis, North Dakota \\
    &&& South Dakota, Oregon \\
  \bottomrule
\end{tabular}
\end{table}

\subsection{Text Detection Model}
PSENet is a segmentation-based model that utilizes CNN features from multiple layers of the network \cite{li2018shape}. It can detect text instances with arbitrary shape or rotation. Given an input image, it first uses an FPN-based Network \cite{lin2017feature} with ResNet \cite{he2016deep} as backbone, then concatenate low-level features with high-level semantic features. The network produces $n$ feature maps of different resolutions $S_1, S_2, ..., S_n$, where each $S_i$ is one segmentation mask that highlights the text instances at a certain scale. Among these masks, $S_1$ gives the segmentation result for the text instances with the smallest scale. After obtaining these segmentation masks, it uses a progressive scale-expansion algorithm to gradually expand all the instances’ kernels to their complete shapes, then obtains the final detection results which are the bounding quadrangles of the detected text instances.

\subsection{Evaluation Metrics}

We use Wolf's metrics\cite{wolf2006object} for evaluation. In the David Rumsey Maps, the annotation is split into multiple polygons if a location name has multiple words or it is in arbitrary shape, while models like PSENet\cite{li2018shape} may detect the entire text instead of splitting it, which should not be penalized as an incorrect detection. The advantage of Wolf's evaluation metric is that it can deal with one to many (one ground truth, many detection polygons) and many to one (many ground truths, one detection polygons) matching.

Let $G$ denote the ground-truth, and $D$ denote the detected polygons. We construct two matrices $\sigma$ and $\tau$. The rows $i = 1, . . . , |G|$ of the matrices correspond to the ground truth polygons and the columns $j = 1, . . . , |D|$ correspond to the detected polygons. The values of the two matrices correspond to area recall and area precision between the row polygon $G_i$ and the column polygon $D_j$:
\begin{align}
    \sigma_{ij} = R_{AR}(G_i,D_j) = \frac{Area(G_i \cap D_j)}{Area(G_i)}
\end{align}
\begin{align}
    \tau_{ij} = P_{AR}(G_i,D_j) = \frac{Area(G_i \cap D_j)}{Area(D_j)}
\end{align}

Two polygons from the two sets $G$ and $D$ are matched only if the overlap ratio for precision and recall are higher than the respective threshold:
\begin{align} 
   \label{eq:wolf_recall_thres}
   \sigma_{ij} \ge t_r \\
   \label{eq:wolf_preci_thres}
   \tau_{ij} \ge t_p
\end{align}

where $ t_r \in [0, 1] $ is the threshold on area recall and $ t_p \in [0, 1] $ is the threshold on area precision, both are set to be $0.5$ in our experiments.

There are three types of matchings:

\begin{description}
   \item[One-to-one matching:] One ground truth polygon $G_i$ matches
one predicted polygon $D_j$ if row $i$ of both matrices contain
only one element satisfying (\ref{eq:wolf_recall_thres}) and (\ref{eq:wolf_preci_thres}) and column $j$ of both matrices contain only one element satisfying (\ref{eq:wolf_recall_thres}) and (\ref{eq:wolf_preci_thres}). 
\item[one-to-many matching (splits):] one ground truth polygon
$G_i$ matches a set $S_o$ of predicted polygon $D_j , j \in
S_o$ if: a sufficiently large proportion of the ground truth
polygon has been detected (condition (\ref{eq:wolf_recall_thres}) in a “scattered” way): $\sum_{j \in S_o} \sigma_{ij} \ge t_r$ and each contributing predicted polygon overlaps enough with the ground truth polygon to be considered a part of it (condition (\ref{eq:wolf_preci_thres}) in a “scattered” way): $\forall j \in S_o : \tau_{ij} \ge t_p$.
\item[many-to-one matching (merges):] one predicted polygon $D_j$ matches against a set $S_m$ of ground truth polygons if: A sufficiently large portion of each ground truth polygon is detected (condition (\ref{eq:wolf_recall_thres}) in a “scattered” version):$\forall i \in S_m : \sigma_{ij} \ge t_r$ and each ground truth polygon has been detected with enough area precision (condition (\ref{eq:wolf_preci_thres}) in a “scattered”
way): $\sum_{i \in S_m} \tau_{ij} \ge t_p$
\end{description}

Based on this matching strategy, the recall and precision measures can be finally defined as follows:

\begin{align} 
   \label{eq:wolf_finall_recall}
    R_{OB}(G,D,t_r,t_p) = \frac{\sum_i Match_G(G_i,D,t_r,t_p)}{|G|}
\end{align}
\begin{align} 
   \label{eq:wolf_finall_precision}
    P_{OB}(G,D,t_r,t_p) = \frac{\sum_j Match_D(D_j,G,t_r,t_p)}{|D|}
\end{align}

where 

\begin{align}
    Match_G(G_i,D,t_r,t_p) = 
    \begin{cases}
        1 & \mbox{$G_i$ matches one detected polygon}  \\
        0 & \mbox{$G_i$ matches no detected polygon} \\
        k & \mbox{$G_i$ matches several detected polygons}
    \end{cases}
\end{align}

\begin{align}
    Match_D(D_j,G,t_r,t_p) = 
    \begin{cases}
        1 & \mbox{$D_j$ matches one truth polygon}  \\
        0 & \mbox{$D_j$ matches no truth polygon} \\
        k & \mbox{$D_j$ matches several truth polygons}
    \end{cases}
\end{align}

where $k \in [0,1]$ is a hyper parameter considered as a penalty for not being a one-to-one match. 

In our experiments, we set $t_r = 0.5, t_p = 0.5$ the same as Wolf et al. \cite{wolf2006object} and $k = 1$, which means that we consider many-to-one and one-to-many matching the same as the one-to-one without penalty.

\subsection{Text Detection Result and Analysis}

\subsubsection{Settings}
We experiment on a state-of-the-art text detection model PSENet\cite{wang2019shape} and report the scores for the three following settings. Notice that in all of these settings, no real historical map images were used for training. 

\begin{itemize}
\item {\verb|ICDAR|}: Model trained on the out-of-domain dataset ICDAR 2015 Incidental Scene Text dataset 
\item {\verb|SynthMap|}: Model trained on our synthetic dataset
\item {\verb|ICDAR+SynthMap|}:  Model first trained on the out-of-domain dataset then fine-tune on our synthetic dataset
\end{itemize}

For all of the above settings, we use exactly same network backbone, ResNet50 \cite{he2016deep}. The backbone weights are initialized from ImageNet \cite{deng2009imagenet}. The three settings only differ on the training set and the training strategy. 

In the \verb|ICDAR| setting, we download the pretrained PSENet weights from the official website and test the model directly on the David Rumsey dataset without further adaptation. The model was trained on the ICDAR15 dataset with image short side resized to 736 during training. It is reported to have 78.5\% F1 score on the ICDAR15 test split. We show the performance of this model on the David Rumsey dataset in the first three columns of Table \ref{tab:pse_main}. The last row computes the average precision, recall and F1 on all the images instead of on the average of the map series. Table \ref{tab:weinman_dataset} records the number of maps in each series and the average number of text labels in each map sheet.

\begin{table*}
\caption{\label{tab:pse_main} PSENet performance on David Rumsey dataset with weights trained on ICDAR, SynthMap and ICDAR+SynthMap. }
\begin{tabular}{l|ccc|ccc|ccc}
\toprule
\multirow{2}{*}{} & \multicolumn{3}{c|}{ICDAR2015} & \multicolumn{3}{c|}{SynthMap} & \multicolumn{3}{c}{ICDAR + SynthMap} \\ \cline{2-10} 
                             & prec.  & recall    & F1        & prec.  & recall      & F1         & prec.    & recall   & F1      \\ \midrule
 D0006  & 84.30\%  &	44.80\%  &	58.50\%  &	68.90\%  &	25.30\%  &	37.00\%  &	79.60\%  &	27.70\%  &	41.10\% \\ \hline
 D0017  & 81.10\%  &	49.30\%  &		\cellcolor{gray!25} 61.30\%  &	85.30\%  &	63.40\%  & 	\cellcolor{gray!25}	72.70\%  &	88.90\%  &	60.80\%  &	\cellcolor{gray!25} 72.20\% \\ \hline
 D0041  & 71.90\%  &	48.90\%  &		\cellcolor{gray!25} 58.20\%  &	71.70\%  &	70.80\%  &	\cellcolor{gray!25}	71.20\%  &	74.90\%  &	72.35\%  &\cellcolor{gray!25} 	73.60\% \\ \hline
 D0042  & 81.28\%  &	34.18\%  &	\cellcolor{gray!25} 47.75\%  &	75.88\%  &	48.65\%  &	\cellcolor{gray!25}	58.83\%  &	77.86\%  &	55.75\%  &	\cellcolor{gray!25} 64.39\% \\ \hline
 D0079  & 45.30\%  &	4.20\%  &	\cellcolor{gray!25} 	7.70\%  &	40.50\%  &	20.60\%  &		\cellcolor{gray!25}27.30\%  &	31.30\%  &	13.60\%  &	\cellcolor{gray!25} 19.00\% \\ \hline
 D0089  & 83.10\%  &	49.90\%  &	62.40\%  &	75.90\%  &	44.60\%  &	56.20\%  &	69.30\%  &	48.10\%  &	56.80\% \\ \hline
 D0090  & 82.80\%  &	55.80\%  &	\cellcolor{gray!25} 66.70\%  &	90.60\%  &	63.40\%  &	\cellcolor{gray!25}	74.60\%  &	91.00\%  &	70.30\%  &\cellcolor{gray!25} 	79.30\% \\ \hline
 D0117  & 89.75\%  &	56.13\%  &		68.90\%  &	72.72\%  &	55.55\%  &	62.95\%  &	82.40\%  &	55.12\%  &	66.02\% \\ \hline
 D5005  & 88.55\%  &	57.68\%  &		\cellcolor{gray!25} 69.57\%  &	78.38\%  &	54.60\%  &	64.23\%  &	82.55\%  &	57.95\%  &\cellcolor{gray!25} 	67.87\% \\ \hline
\textbf{All} & \textbf{82.76}\% &	\textbf{45.00}\% &	\textbf{57.32}\% &	\textbf{74.90}\% &	\textbf{51.73}\% &	\textbf{60.62}\% &	\textbf{78.51}\% &	\textbf{55.25}\% &	\textbf{64.25}\% \\ 
\bottomrule
\end{tabular}
\end{table*}

\begin{table*}
\caption{\label{tab:pse_tptr} PSENet (ICDAR+SynthMap) $F1$ scores with varying $t_p$ and $t_r$ threshholds on one sample map image}
\begin{tabular}{c|ccc|ccc|ccc}
\toprule
& $t_p$=0.1 	& 0.2 	& 0.3 	& 0.4 	& 0.5 	& 0.6 	& 0.7 	& 0.8	& 0.9 \\ \midrule
$t_r$=0.1 & 78.51\% 	& 77.95\% 	& 77.67\% 	& 74.42\% 	& 70.48\% 	& 65.38\% 	& 59.92\% 	& 47.90\% 	& 35.69\% \\ \hline
0.2 & 77.28\% 	& 76.31\% 	& 75.31\% 	& 71.68\% 	& 67.87\% 	& 61.47\% 	& 55.93\% 	& 44.57\% 	& 32.87\% \\ \hline
0.3 & 75.87\% 	& 74.80\% 	& 73.09\% 	& 69.38\% 	& 65.78\% 	& 58.65\% 	& 52.64\% 	& 42.29\% 	& 30.13\% \\ \hline
0.4 & 72.85\% 	& 71.85\% 	& 70.00\% 	& 66.74\% 	& 61.94\% 	& 54.35\% 	& 49.92\% 	& 39.23\% 	& 28.51\% \\ \hline
0.5 & 70.93\% 	& 69.93\% 	& 67.63\% 	& 63.79\% 	& 56.79\% 	& 50.39\% 	& 46.01\% 	& 35.79\% 	& 26.39\% \\ \hline
0.6 & 69.66\% 	& 68.25\% 	& 65.57\% 	& 60.63\% 	& 53.05\% 	& 47.18\% 	& 42.44\% 	& 32.87\% 	& 23.98\% \\ \hline
0.7 & 66.19\% 	& 64.64\% 	& 61.34\% 	& 56.03\% 	& 49.11\% 	& 44.20\% 	& 37.97\% 	& 29.24\% 	& 20.36\% \\ \hline
0.8 & 61.06\% 	& 58.57\% 	& 53.38\% 	& 46.86\% 	& 38.45\% 	& 33.85\% 	& 28.11\% 	& 22.48\% 	& 15.69\% \\ \hline
0.9 & 51.15\% 	& 43.47\% 	& 37.60\% 	& 28.14\% 	& 22.96\% 	& 19.05\% 	& 16.63\% 	& 14.16\% 	& 10.85\% \\
\bottomrule
\end{tabular}
\end{table*}

\begin{figure*}[h]
\begin{subfigure}{\textwidth}
\frame{\includegraphics[width=\textwidth]{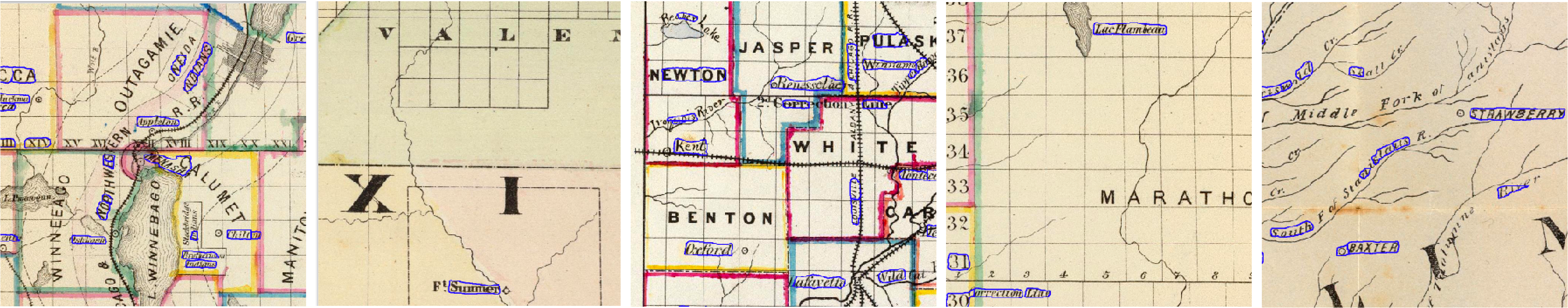}}
    \caption{PSENet trained with ICDAR 2015}
\end{subfigure}
\hfill
\begin{subfigure}{\textwidth}
 \frame{\includegraphics[width=\textwidth]{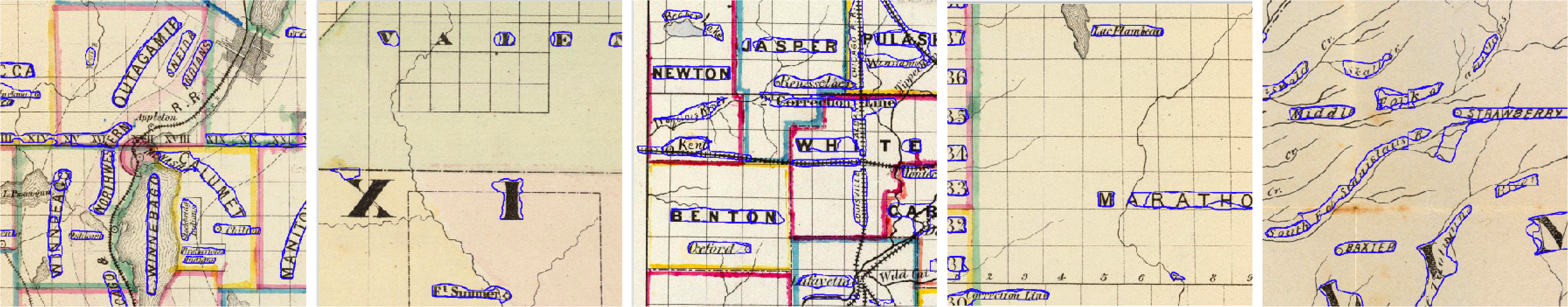}}
    \caption{PSENet trained with our SynthMap}
\end{subfigure}
\hfill
\begin{subfigure}{\textwidth}
 \frame{\includegraphics[width=\textwidth]{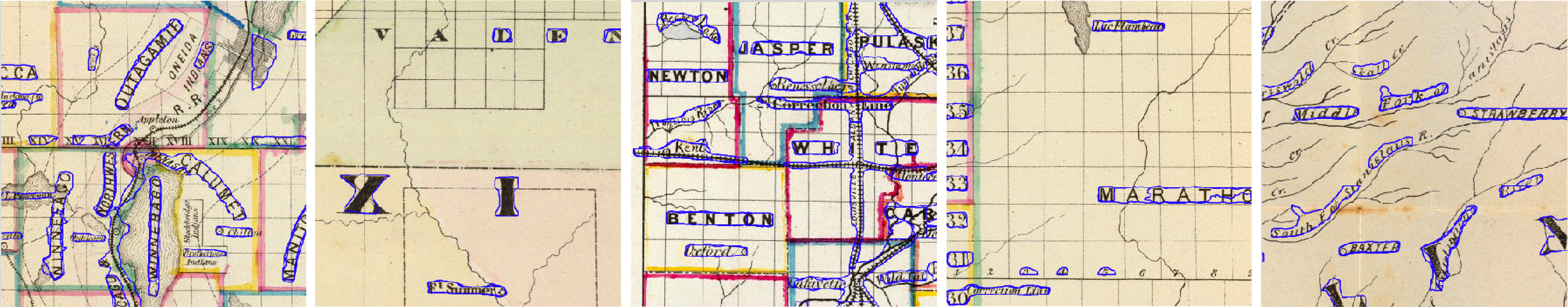}}
    \caption{PSENet trained with ICDAR then finetune on our SynthMap}
\end{subfigure}
\caption{Qualitative result comparison. }
\label{fig:qualitative_compare}
\end{figure*}

In the \verb|SynthMap| setting, we train the PSENet model from scratch. To enlarge the color variance of the SynthMap training set, we created some images with SynthMap text layers but with OSM (no-text) background. In the text layer, we added Gaussian noise to the text region to simulate the "worn-out" effect in the historical map images. During training, in addition to the regular data augmentation techniques such as flipping, resizing, and cropping, we also added ColorJittering augmentation with hue and contrast change. The base contrast of the original image is 1, and the augmentation randomly changes the contrast of the image within the range [0.5,1.5]. For reference, the valid contrast values range is [0,2]. Contrast value 0 gives a solid gray image, and 2 increases the contrast by a factor of 2. The base hue value is 0, and we randomly adjust the hue values within range [-0.5,0.5]. This augmentation shifts the hue value in the HSV space, where 0.5 and -0.5 give completely reversal hues for the image. 

In the \verb|ICDAR+SynthMap| setting, we first load the weights trained on the ICDAR dataset, then fine-tune the model with SynthMap images. We use the same training strategy as in the second setting: \verb|SynthMap|. Using ICDAR-pretrained weights can be seen as the case with better weight initialization.

\subsubsection{Analysis}

Table \ref{tab:pse_main} summarizes the quantitative results for the three settings. We can observe that the PSENet model trained on SynthMap from scratch performs better than that trained on the ICDAR dataset. This is likely due to the fact that the synthetic map images are closer to the domain of map images while the scene images from ICDAR are quite different from the map images. When the PSENet model fine-tunes the weights on ICDAR with SynthMap images, the accuracy boosted even further from 57.32\% to 64.25\%. The improvement in the F1 score is mostly due to the improvement on recall. The average recall rate for all the map images increased from 45.00\% to 55.25\% . 
   
From Table \ref{tab:pse_main}, we can also see that the \verb|ICDAR+SynthMap| setting improved on almost all the map series except for D006, D0089, and D0117. We thus visualize the text detection results where the PSENet benefits from SynthMap dataset in Figure \ref{fig:qualitative_compare}. The figure shows that the model trained with \verb|ICDAR| settings fails to detect many of the curved text regions, and it performs badly for the horizontal text labels. Another hard case is when the font size is very large, or the characters are very widely separated. PSENet trained with  \verb|ICDAR+SynthMap| sometimes still suffers from the widely-separated text labels, but the performance for this case still improved quite a bit. For the large font, PSENet with  \verb|ICDAR-SynthMap| draws tight polygons around the edge of the character, and this sometimes might cause the recall $\sigma_{ij}$ to drop lower than 0.5 and thus be considered as a false negative. So if we loosen the $t_r$ threshold for $\sigma_{ij}$, the accuracy would be further increased. Table \ref{tab:pse_tptr} shows the performance with varying $t_p$ and $t_r$ thresholds. Smaller values in the thresholds lead to higher numbers in the $F1$ score. 

\begin{figure}[h]
\includegraphics[width=\linewidth]{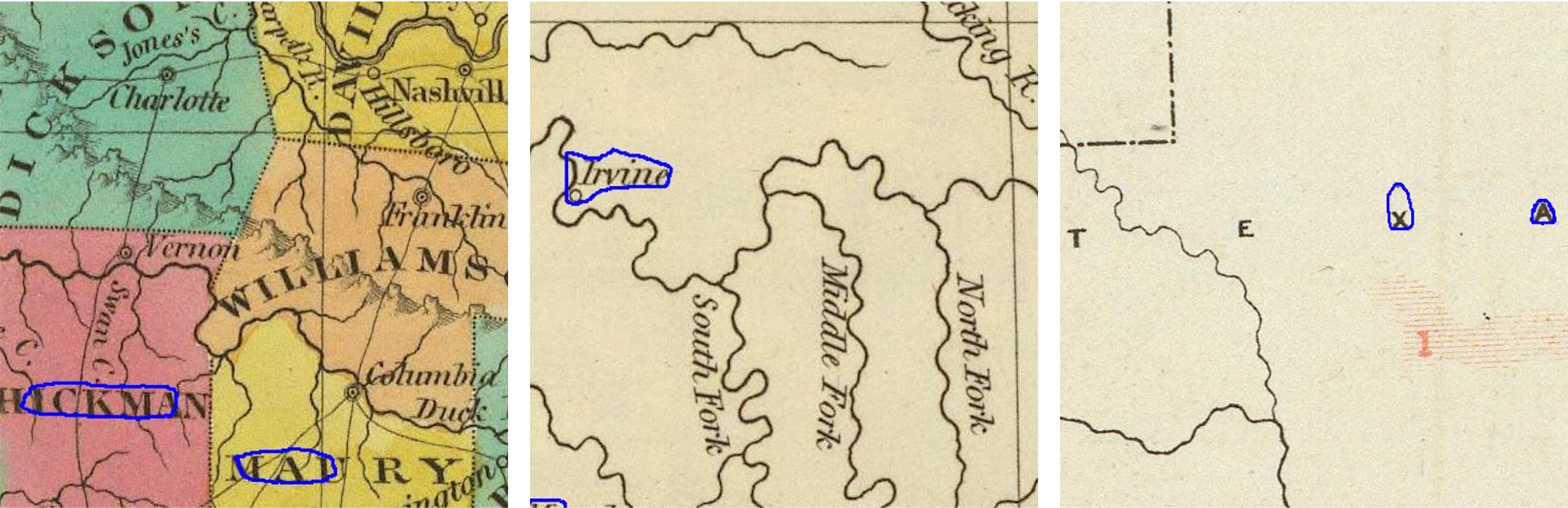}
\caption{This figure shows the failure cases when trained on ICDAR and fine-tuned on SynthMap. \textit{Left}: The model fails with unseen background color. \textit{Middle}: The model fails to detect vertical text regions. \textit{Right}: Text detection fails when the characters of the text labels are very widely separated. }
\label{fig:failure}
\end{figure}

In Figure \ref{fig:failure}, we show several sample images where the PSENet model ( \verb|ICDAR+SynthMap|) fails. This gives us an idea on the scenarios where our SynthMap dataset has not covered well yet. When the background has large areas where the color has not appeared in the training set before, the model fails to detect text regions on such backgrounds. Also, the model does not perform well on vertical text regions and widely separated text labels.

\section{Related Work}
\subsection{Text Detection Datasets}
There are many text detection datasets collected in different domains, such as scene images, video frames, and research publications \cite{karatzas2015icdar,yao2012detecting,veit2016coco, biopublication2017, rctw17}. The International Conference on Document Analysis and Recognition (ICDAR) has made a great effort on organizing text detection competitions \cite{karatzas2015icdar, biopublication2017} and encourages the development of datasets and algorithms. During 2013-2015, the competitions focused on born-digital documents and focused scene text images. After 2015, incidental scene text detection has attracted more attention. Scene images were taken by various devices (e.g., pocket cameras, cellphones, and drones) and collected to increase the variety of the datasets. Text detection was no longer restricted to English, a multi-lingual text detection dataset was also created by ICDAR in 2019 \cite{icdar2019mlt}. 

MSRA-TD500 \cite{yao2012detecting} is another dataset of scene image text detection. It contains 500 images for both indoor and outdoor scenes. Although the size of the dataset is relatively small, the images have large variations in the background lighting condition, font size, style, and image resolution. The number of images in the ICDAR datasets and COCO-Text MSRA-TD500 are comparably small for training deep learning models. It is common to first pretrain the model on some large-scale datasets then fine-tune on one of the previous datasets. 

COCO-Text \cite{veit2016coco} is a much larger dataset that contain 63,686 images with 145,859 text instances. It covers both machine-printed and handwritten text in different languages. 

Aside from those multi-lingual datasets \cite{veit2016coco, icdar2019mlt}, there are some datasets for Chinese character detection only. RCTW-17 \cite{rctw17} and Chinese Text in the wild  \cite{yuan2018chinese} are two benchmark datasets for this purpose. RCTW-17 includes more than 12,000 images taken by either phone cameras or phone screenshots. The images cover both indoor and outdoor scenes, including street views, menus, and posters. The text labels are annotated with quadrilaterals following the ICDAR 2015 \cite{karatzas2015icdar} convention. Chinese Text in the wild \cite{yuan2018chinese} is an even larger dataset that contains 32,286 street view images with about 1 million Chinese characters. There are 3,850 unique characters that are commonly used in real-life scenarios. The dataset has a large diversity in text font size, style, shape, and occlusion. 

All the datasets mentioned above do not contain (or are not able to annotate) the curved text labels. Thus two other datasets are proposed for the curved-text detection: SCUT-CTW1500 \cite{scut-ctw1500} and TotalText \cite{totaltext}. SCUT-CTW1500 includes 1,500 images with 10,751 text labels. 3,530 are curved text instances among all the text labels. The images are collected from various sources such as web pages, image libraries, and phone cameras. The images have both English words and Chinese characters, and many of those are multi-oriented. TotalText \cite{totaltext} is roughly of the same size as SCUT-CTW1500, and it contains 1,555 images that have text labels in different orientations and shapes.

The above datasets are mainly for the scene text detection, and text detection datasets in the historical map domain are pretty rare. The David Rumsey Maps dataset \cite{weinman2017geographic} is one valuable historical text detection dataset annotated by Weinman et al. This is the dataset that we use in this paper for evaluation. 


\subsection{Synthetic Data Generation}
The data collection and annotation require a lot of manual work, and some researchers have proposed creating synthetic datasets for the text detection tasks. SynthText \cite{synthtext} is a very large scale dataset with about 800,000 real scene images and about 8 million synthetic text instances. Each text label has character level, word level, and bounding-boxes level annotations. SynthText uses a segmentation-based method to find reasonable areas for label placement, such that the resulting synthetic images look very natural. UnrealText \cite{long2020unrealtext} contains about 600K synthetic images with about 12 million word instances. It utilizes the UnrealText 3D graphics engine to place the text lables on valid 3D object surfaces to achieve a realistic appearance. 

The motivation of our proposed method and the above two papers are very similar. We rely on synthetic data generation to produce a large (potentially unlimited) amount of annotated data. In contract to SynthText and UnrealText, our proposed method generates the text data in the historical map domain and supports the annotation of arbitary shaped and oriented text labels.

\section{Conclusion and Future Work}
This paper presented an end-to-end pipeline, SynthMap, to generate an unlimited amount of synthetic historical map images from OpenStreetMap (OSM). SynthMap first uses a style transfer network to convert OSM tiles to the NLS historical map style. Then SynthMap uses the QGIS PAL API to place the text labels on the synthetic map layers. We propose an annotation generation algorithm to automatically generate polygon, centerline, and local height information to represent the text label boundaries. With this method, we created a SynthMap dataset with more than 10K synthetic historical map images. The data can be used as the training data for the map text detection tasks. We adopted a state-of-the-art text detection model PSENet and train the model with our SynthMap dataset. We compared the performance of the model when trained on the out-of-domain dataset and observe a large improvement in the text detection accuracy. The proposed method is a general pipeline,  not restricted to the CycleGAN model for style transfer. CycleGAN can be replaced with any other more advanced style transfer models in the future to generate synthetic map images with higher quality. SynthMap can also potentially generate a large amount of training data for other map analysis tasks, such as word-linking and road delineation.

\begin{acks}
This material is based upon work supported in part by the National Science Foundation under Grant Nos. IIS 1564164 (to the University of
Southern California) and IIS 1563933 (to the University of Colorado at Boulder), NVIDIA Corporation,  the National Endowment for the Humanities under Award No. HC-278125-21, and the University of Minnesota, Computer Science \& Engineering Faculty startup funds.

\end{acks}

\bibliographystyle{ACM-Reference-Format}
\bibliography{reference}

\end{document}